\documentclass[conference]{IEEEtran}
\IEEEoverridecommandlockouts
\usepackage{cite}
\usepackage{amsmath,amssymb,amsfonts}
\usepackage{algorithmic}
\usepackage{graphicx}
\usepackage{textcomp}
\usepackage{xcolor}
\def\BibTeX{{\rm B\kern-.05em{\sc i\kern-.025em b}\kern-.08em
    T\kern-.1667em\lower.7ex\hbox{E}\kern-.125emX}}
\begin{document}

\title{Audio-Driven Talking Face Generation with Blink Embedding and Hash Grid Landmarks Encoding\\
}

\author{\IEEEauthorblockN{Yuhui Zhang$^1$, Hui Yu$^{2*}$, Wei Liang$^{1,2}$, Sunjie, Zhang$^1$}
\IEEEauthorblockA{\textit{$^1$ Department of Control Science and Engineering, University of Shanghai for Science and Technology, Shanghai 200093, China}\\
 \textit{$^2$ School of Psychology and Neuroscience, University of Glasgow, 62 Hillhead Street, Glasgow G12 8QB, Scotland, UK} \\
 \thanks{$^*$Corresponding author: Hui Yu, Email: hui.yu@glasgow.ac.uk}}
}

\maketitle

\begin{abstract}
Dynamic Neural Radiance Fields (NeRF) have demonstrated considerable success in generating high-fidelity 3D models of talking portraits. Despite significant advancements in the rendering speed and generation quality, challenges persist in accurately and efficiently capturing mouth movements in talking portraits. To tackle this challenge, we propose an automatic method based on blink embedding and hash grid landmarks encoding in this study, which can substantially enhance the fidelity of talking faces.  Specifically, we leverage facial features encoded as conditional features and integrate audio features as residual terms into our model through a Dynamic Landmark Transformer. Furthermore, we employ neural radiance fields to model the entire face, resulting in a lifelike face representation. Experimental evaluations have validated the superiority of our approach to existing methods.

\end{abstract}


\section{Introduction}
Over the past decade, there has been impressive advancement in Digital Human technologies, particularly in the emergence of the meta-universe applications. Within this paradigm, 3D face generation technologies have garnered increasing attention and application across various research fields. One of the paramount challenges in the realm of Digital Human and metaverse development lies in the generation of speaker portraits from arbitrary speech audio. In earlier works, conventional methods for creating talking faces predominantly leveraged Generative Adversarial Networks (GANs) or image-to-image transformation techniques as image renderers\cite{zhou2019talking}\cite{zhou2020makelttalk}. Some approaches also integrated intermediate parameters such as 2D landmarks to facilitate image transformation\cite{prajwal2020wav2liplip}, exemplified by additional processing like wav2lip. However, these methodologies often encounter limitations in image quality due to their reliance on 2D geometric modeling, consequently yielding suboptimal outcomes.

In response to these challenges, the field has seen a shift towards more sophisticated approaches that aim to enhance image fidelity and realism. These advancements entail the exploration of novel algorithms and technologies capable of leveraging 3D facial representations and audiovisual synchronization techniques to generate speaker portraits with greater accuracy and realism. By moving beyond the constraints of 2D geometric modeling, these cutting-edge methodologies attempt to overcome previous limitations and make progress for more immersive and lifelike digital human experiences within the metaverse application and beyond.

Subsequently, a series of methods based on three-dimensional models have been proposed to generate speaker portraits utilizing Three-Dimensional Morphable Models (3DMM)\cite{blanz20033dmm}. Leveraging the advantages of three-dimensional structural modeling, these methods achieved a more natural speaking style compared to two-dimensional approaches. Recently, Neural Radiance Fields (NeRF) technology\cite{mildenhall2021nerf} was proposed to reconstruct complex 3D scenes. it has gradually been introduced into the generation of speaking faces. However, techniques like AD-NeRF\cite{guo2021AD-nerf} and DFRF\cite{shen2022DFRF} still relied on end-to-end audio-to-video NeRF renderer approaches, resulting in excessively long generation times for target individuals, far from achieving real-time rendering effects.

Rad-NeRF\cite{tang2022RAD-nerf} further integrated instant-NGP\cite{muller2022instant-ngp} into the reconstruction of speaking individual portraits, achieving significant efficiency improvements, albeit still scarified in image quality. The work of Geneface\cite{ye2023geneface}\cite{ye2023geneface++} incorporated audio-driven 3D landmarks into the NeRF framework, enhancing its generalization capabilities to out-of-domain audio. ER-NeRF\cite{li2023ER-nerf}, building upon RAD-NeRF, enhanced the representation efficiency of the neural radiance field by utilizing three plane hashing encoders, highly accelerating the rendering speed. However, it still exhibits some unsatisfying phenomena such as partial frames with fog-like artifacts\cite{li2023ER-nerf}.

NeRF, as a neural rendering technique, utilizes five-dimensional inputs to capture intricate 3D surfaces\cite{mildenhall2021nerf}. Initially, it was developed for rendering static bounded fields, but it has evolved to support dynamic and unbounded scenes with advancements in technology. Subsequently, NeRF was adapted for constructing talking faces, yielding promising results. However, the original NeRF suffers from limitations in the rendering speed and mouth accuracy, and its applicability is also constrained to custom-designed speakers due to technological constraints. Presently, approaches employing neural radiance fields for face reconstruction can be broadly categorized into two categories: implicit representations that incorporate audio information to deform facial structure space\cite{shen2022DFRF}\cite{li2023ER-nerf}\cite{wang2023cp-eb}, and explicit methods utilizing 3D landmarks or facial expression parameters\cite{ye2023geneface}\cite{ye2023geneface++}. These methodologies encounter common challenges, including enhancing mouth and facial morphology representation, and effectively capturing audio-driven mouth dynamics\cite{su2023dt-NERF}.

To address these challenges, we propose a coarse to fine method in this study. In the first stage,  we first encode the input audio to extract the audio information through the DeepSpeech module, and then feed the output audio features into the Dynamic Landmark Transformer motion generator to transform the audio into the corresponding 3D face landmark.
 In the second stage, we integrate a blink prediction module in the process to alleviate the uncanny valley effect, which can further enrich the dynamic facial features. Furthermore, leveraging the hash grid from instant-NGP\cite{muller2022instant-ngp}, we perform positional encoding to input the extracted features of the corresponding 3D landmarks into the neural radiance field. This significantly enhances rendering speed, achieving real-time optimization. Moreover, it enriches the density and color grids within NeRF, facilitating smoother transitions of the face from canonical to dynamic spaces.\cite{9406832}\cite{YU2012152}

\section{ Related Work}

\subsection{2D-Based Talking-Head Synthesis}

The earliest works in this domain typically fall under 2D-based methods, utilizing techniques such as GAN or image-to-image conversion as the image renderer\cite{zhang2020davd}\cite{zhou2020makelttalk}\cite{zhou2019talking}. Some research studies employed intermediate parameters like 2D landmarks to facilitate image transformation, i.e. as part of wav2lip\cite{prajwal2020wav2liplip}. While these pixel-based facial reconstruction methods produce images with a good quality, they suffer from precise constraints for audio-visual synchronization.

In StyleHEAT\cite{yin2022styleheat}, for instance, the authors demonstrated the utilization of StyleGAN\cite{abdal2019image2stylegan} models to create speaking faces guided by speech embeddings but controlled intuitively or via attribute editing. Audio2head achieved single-shot speaking head generation by predicting flow-based motion fields\cite{wang2021audio2head}. Additionally, schemes utilizing diffusion models, such as difftalk\cite{shen2023difftalk}, employed conditional diffusion models for high-quality generic speaking head synthesis. To achieve personality-aware generalized synthesis, dual-reference images were further utilized as conditions. This approach notably outperformed other 2D-based methods in generated image quality.

However, the aforementioned 2D-based speaking face approaches are constrained to modeling on a two-dimensional plane, so they are unable to capture the three-dimensional characteristics of speaking heads effectively. It thus challenging for those methods to generate vivid and natural three-dimensional speaking faces.

\subsection{3D-Based Talking-Head Synthesis}

Synthesis of speaking heads based on three-dimensional models: A large body of works based on three-dimensional models utilize Three-Dimensional Morphable Models (3DMM) to generate speaking head images. Leveraging the advantages of three-dimensional structural modeling, this type of methods can achieve a more natural speaking style compared to two-dimensional approaches. For example, Thies, J. \cite{zollhofer2018state}, could generate realistic and natural talking head videos. In\cite{ren2021pirenderer}, a Portrait Image Renderer (PIRenderer) was introduced, utilizing parameters of a three-dimensional deformable face model to control facial movements. In \cite{pham2017speech}, an implicit emotion displacement learner was employed alongside dense deformation fields to attain high-quality images. However, since the network's optimization was designed for identity-specific learning, it required training for each identity on large datasets. Another common limitation arises from the loss of information incurred by using intermediate 3DMM\cite{blanz20033dmm} parameters, leading to the loss of facial details in the generated faces.

\subsection{Neural Radiance Fields for Rendering Face}

Neural Radiance Fields (NeRF) represent a deep learning model tailored for three-dimensional implicit space modeling\cite{mildenhall2021nerf}. It captures scenes under known viewpoints without the need for intermediate 3D reconstruction processes. By synthesizing realistic three-dimensional scenes solely based on pose intrinsics and images, NERF has found extensive applications and laid the groundwork for various research areas.
Initially, Guo et al. \cite{guo2021AD-nerf} proposed AD-NERF, which used audio-conditioned implicit functions to generate dynamic neural radiance fields. It employed volume rendering to synthesize high-fidelity talking head videos corresponding to audio signals. Early methods implicitly constructed head models, leading to trade-offs between speed, efficiency, and practicality. In order to achieve training with minimal data, DFRF adjusted facial radiance fields on 2D reference images to learn facial priors, significantly reducing the required data scale (only tens of seconds of video) while improving convergence speed\cite{shen2022DFRF}.

With the evolution of NeRF technologies, RAD-NeRF\cite{tang2022RAD-nerf} introduced instant-NGP\cite{muller2022instant-ngp} into the reconstruction of speaking individual portraits, leading to notable efficiency improvements. The work of Geneface\cite{ye2023geneface}\cite{ye2023geneface++} integrated audio-driven 3D landmarks into NeRF, enhancing its generalization capabilities to out-of-domain audio. And ER-NeRF, by utilizing three plane hashing encoders to prune empty spatial regions, improved the representation efficiency of neural radiance fields\cite{li2023ER-nerf}.
These advancements demonstrate the continuous refinement and enhancement of NeRF-based approaches in synthesizing realistic three-dimensional scenes, contributing 
\begin{figure*}[htbp]
\centering
\includegraphics[width=0.95\textwidth]{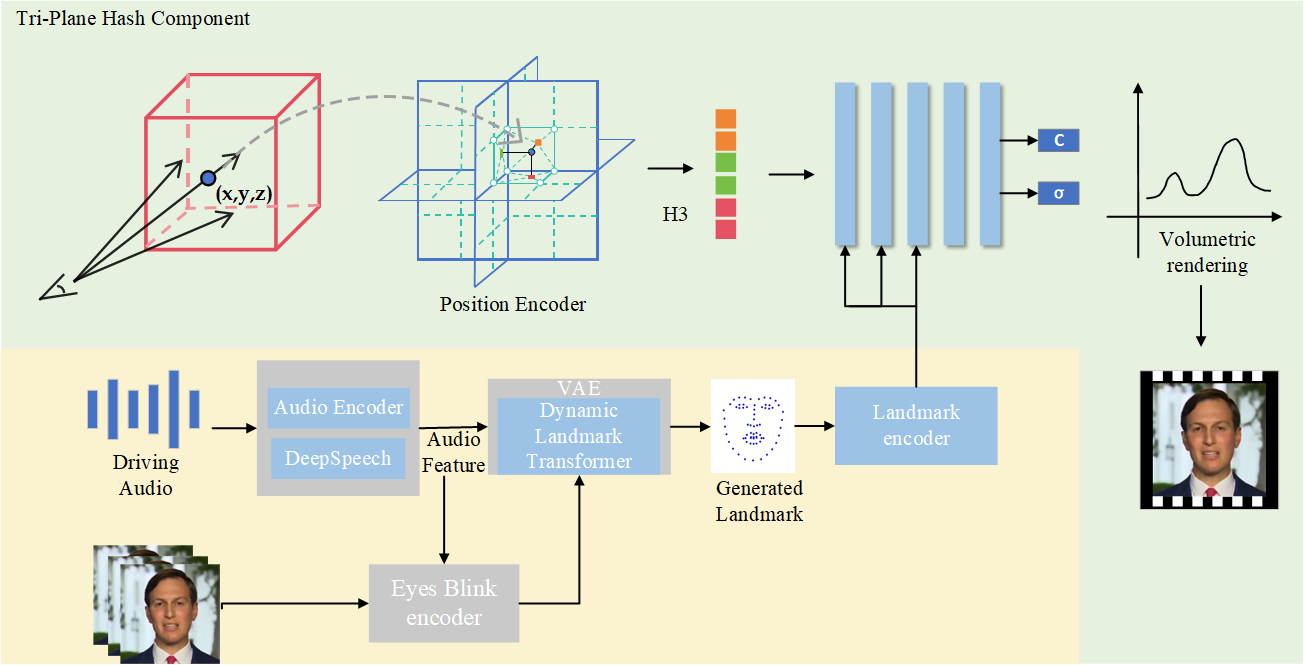}
\caption{ network architecture Our model employs tri-plana hashing to represent facial regions and a Dynamic landmark Transformer to process audio features as queries to retrieve coordinates and corresponding attributes within NeRF space. Subsequently, we process the volumetric density (denoted as (c, $\sigma$)) of the facial representation from NeRF space by volume rendering to form a facial video of the speaking head.}
\label{fig}
\end{figure*}
to diverse application scenarios and pushing the boundaries of realism in digital content creation.\cite{10153641}\cite{xia2021relation}

\section{Method}

\subsection{Problem Statement}\label{AA}
The NeRF function serves as a representation of a continuous scene, formulated as a function with the inputs consisting of 5-dimensional vectors\cite{mildenhall2021nerf}. These vectors are the 3D coordinate position of a spatial point denoted as $x= \left ( x,y,z\right )$, along with the viewpoint direction denoted as $d= \left ( \varphi ,\theta \right )$. This neural network can be mathematically expressed as follows:
\begin{equation}
\mathcal{F}: \left ( x,d\right )\rightarrow \left ( c,\sigma \right )
\end{equation}

In the output of NeRF, $\sigma$ represents the density of the corresponding 3D position, while $c = (R, G, B)$ denotes the perspective-dependent colour of the 3D point. NeRF adopts the traditional volume rendering method to define a camera ray as $r(t) = o + td$, where $o$ is the origin of the ray, $d$ is the previously mentioned direction of the camera ray, and $tn$ and $tf$ represent the proximal and distal boundaries of $t$, respectively.  The colour of the sampling point through which this ray passes can then be expressed as an integral of:
\begin{equation}
\hat{C}(r)=\int_{t_n}^{t_f} \sigma(\mathbf{r}(t)) \cdot \mathbf{c}(\mathbf{r}(t), \mathbf{d}) \cdot T(t) d t
\end{equation}

where $T(t)$ denotes the accumulated transparency of the ray from tn to t. It represents the probability of the ray traversing without intersecting any particles within this interval, akin to its unobstructed path through the scene. This can be mathematically expressed as:
\begin{equation}
T(t)=\exp \left(-\int_{t_n}^t \sigma(\mathbf{r}(s)) d s\right)
\end{equation}

The Instant-NGP utilizes hash encoding to represent sampling points in space\cite{muller2022instant-ngp}, facilitating efficient neural representations. It adopts a multi-resolution framework for effective querying of colour c and density $\sigma$, thereby reducing the training time. Inspired by Instant-NGP\cite{muller2022instant-ngp}, the ER-NeRF\cite{li2023ER-nerf} framework has been developed to generate speaker facial images efficiently. This framework employs a three-plane hash approach to address a primary challenge encountered when using hash encoding-hash collision.
\begin{equation}
\mathbf{f}_{\mathbf{x}}=\mathcal{H}^{\mathbf{X Y}}(x, y) \oplus \mathcal{H}^{\mathbf{Y Z}}(y, z) \oplus \mathcal{H}^{\mathbf{X Z}}(x, z) 
\end{equation}

where H represents the multi-resolution hash encoder, while $\oplus$ denotes the concatenation operator, appending features into a three-channel vector. Therefore, the input to the final MLP decoder comprises $fx$ , viewing direction $d$ , and audio features A. The implicit function represented by the three-plane hash can be formulated as:
\begin{equation}
\mathcal{F}^{\mathcal{H}}:\left(\mathbf{x}, \mathbf{d}, \mathcal{A} ; \mathcal{H}^3\right) \rightarrow(\mathbf{c}, \sigma)
\end{equation}

\subsection{Dynamic Landmark Radiance Field}

In the task of synthesizing speaker portraits driven by audio, extracting rich and discriminative conditional features from the input-driven audio is crucial. We propose a dynamic landmark generator to accurately and expressively generate facial landmarks from the given input audio.

To achieve better synchronization between audio and lip movements, the dynamic landmarks generated by the audio-predicting motion module are imported into the MLP used for radiance field rendering. Initially, we employ a pre-trained RNN-based speech recognition module, such as DeepSpeech\cite{hannun2014deepspeech}, to extract acoustic features from the input audio signal. To ensure frame-to-frame consistency, a time filtering module is further introduced to compute smoothed acoustic features A \cite{shen2022DFRF}. Subsequently, these extracted audio features are passed through a Variational Autoencoder (VAE) framework to learn the latent representation of the audio data. Within the VAE, these audio features are encoded into distribution parameters in the latent space and then reconstructed to audio signals through the decoder component.

For better representation of head movement, we adopt the 68 Landmarks extracted from the reconstructed 3D head to represent facial motion. Considering the personalized nature of our task, we jointly train the dynamic landmark transformer with the target individual's video, encoded using blink encoding$\left(III.C\right)$, and target audio. During inference, the dynamic landmark transformer is utilized to predict facial landmarks.
\begin{equation}
\mathcal{P}= DLT\left ( A,V\right )
\end{equation}

Let $P$ denote the predicted facial landmark points, and $v$ represent the video encoded using blink encoding. We thus can achieve cross-modal generation from audio signals to facial motion landmark points. In this process, a positional loss is introduced, which is the $L1$ distance between the predicted position and the target position:
\begin{equation}
L_{\mathrm{p}}=\frac{1}{M \times F} \sum_{m=1}^M \sum_{f=1}^T\left\|\hat{p}_f^m-p_f^m\right\|_1 .
\end{equation}
where $p_f^m$ represents the position of landmark point \( m \) in frame \( f \).

\subsection{Facial and Eye Movement Integration Module}
In order to mitigate the uncanny valley effect and achieve realistic blinking in the generated speaking head videos, it is crucial to accurately incorporate eye movements. While the dynamic landmark module effectively captures facial movements in response to audio stimuli, the correlation between audio and eye movements is relatively weak. To address this issue, eye landmarks are separately extracted, which can reduce complexity and intermediary losses while ensuring natural eye movements in the generated speaking head videos.

Ground truth eye movements are extracted using OpenFace\cite{baltruvsaitis2016openface}, a tool capable of capturing facial action units with landmarks from videos. In OpenFace, the intensity of blinking actions is measured on a scale from 0 to 5 and recorded in a csv file.

As shown in Fig. 2, in each audio feature frame, we extract the corresponding Action Units (AUs), combining the temporal features of the entire audio segment with the local features of each frame. These are then mapped into a Facial Action Unit Feature Vector. Subsequently, we import the extracted vectors into the mapping module to obtain their eye conditions. From the 68 facial landmarks, we extract 6 landmarks for each eye region, and predict the next frame's eye state using the Eyes Movement prediction module. Finally, we input the blink features into the dynamic landmark Radiance Field, together serving as the final implicit representation in the latent space.

\begin{figure}[htbp]
\centering
\includegraphics[width=0.5\textwidth]{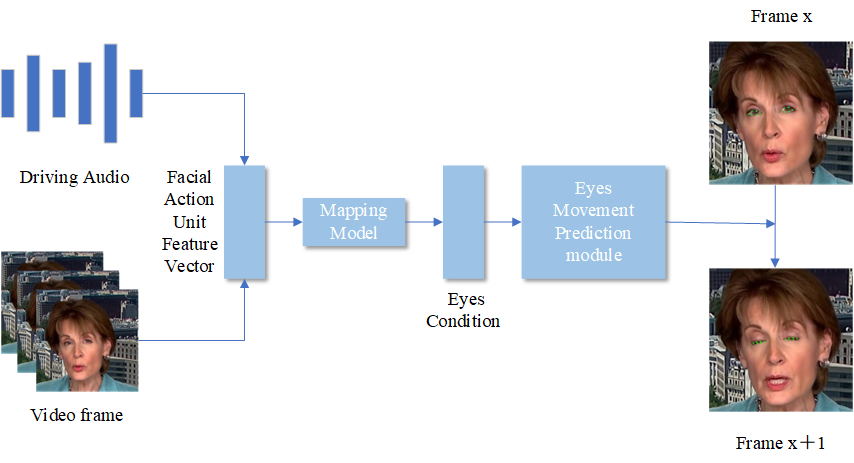}
\caption{The mapping network comprising convolutional layers and fully connected layers, utilizing only six landmarks surrounding the eyes. This network takes as input an eye-related vector and learns the current size of the eyes, which is then fed into a control module. The control module predicts the movement of relevant landmarks for eye transformation.}
\label{fig}
\end{figure}

\subsection{Training Details}
Due to the integration of the triplane-hash and dynamic landmark modules, we employ a two-stage coarse-to-fine training process to achieve a higher image quality. We follow the settings in AD-nerf\cite{guo2021AD-nerf} in the coarse training phase and use pixel-level loss, employing Mean Squared Error (MSE) loss for the colour component of the facial region. Here, $\hat{C}(r)$ represents the predicted colour, c(i) represents the colour of the ground truth.
\begin{equation}
\mathcal{L}_{\text {coarse }}=\sum_{\mathrm{i} \in \mathcal{I}}\|C(\mathrm{i})-\hat{C}(\mathrm{i})\|_2^2
\end{equation}

After the coarse training stage, we proceed to fine-tune the model specifically targeting the mouth region. During the fine training stage, we utilize both Mean Squared Error (MSE) loss and LPIPS (Local Pixelwise Importance-based Perceptual) loss for the mouth region. We randomly sample a set of patches $P$, and the value of $\lambda$ is set to 0.001.
\begin{equation}
\mathcal{L}_{\text {fine }}=\sum_{\mathrm{i} \in \mathcal{P}}\|C(\mathrm{i})-\hat{C}(\mathrm{i})\|_2^2+\lambda \operatorname{LPIPS}(\hat{\mathcal{P}}, \mathcal{P})
\end{equation}

\section{Experiment}
\subsection{Experimental Settings}
We have conducted experiments on several publicly available videos on the internet ranging from 3 to 5 minutes long, with a frame rate of 25 fps and a resolution of 512x512 pixels in DFRF and geneface. We have compared our model with DFRF\cite{shen2022DFRF}, ER-NeRF\cite{li2023ER-nerf}, Geneface\cite{ye2023geneface}, and other models. The model is trained using the Adam optimizer and the PyTorch framework, and all experiments are conducted on a single RTX 4090 GPU.

We have evaluated the performance using a set of quantitative metrics and visualizations. Specifically, we use Peak Signal-to-Noise Ratio $\left(PSNR \uparrow\right)$ and Learned Perceptual Image Patch Similarity $\left(LPIPS \downarrow\right)$ to assess the image quality. PSNR tends to favour blurry images , thus we utilize the representative perceptual metric LPIPS. Additionally, we utilize SyncNet $\left(Sync \right)$
 to measure audio-visual synchronization, where a smaller absolute value indicates better synchronization. Landmark Distance $\left(LMD \downarrow\right)$ serves as a metric to evaluate the quality of facial image generation, measuring the distance between facial landmarks in generated and real images to assess the accuracy of facial features. Finally, we utilize Fréchet Inception Distance $\left(FID \downarrow\right)$ to evaluate the authenticity of generated images. 

\subsection{Quantitative Evaluation}
We employ a self-driven approach to conduct experiments on Emmanuel Macron's video\cite{ye2023geneface} footage. Fig. 3 illustrates a comparative analysis between our method's output driven by its own audio and ground truth. Upon closer inspection, our method demonstrates superior fidelity in capturing finer details, particularly evident in simulating wrinkles and synchronizing lip movements with audio cues.

\begin{figure}[htbp]
\centering
\includegraphics[width=0.5\textwidth]{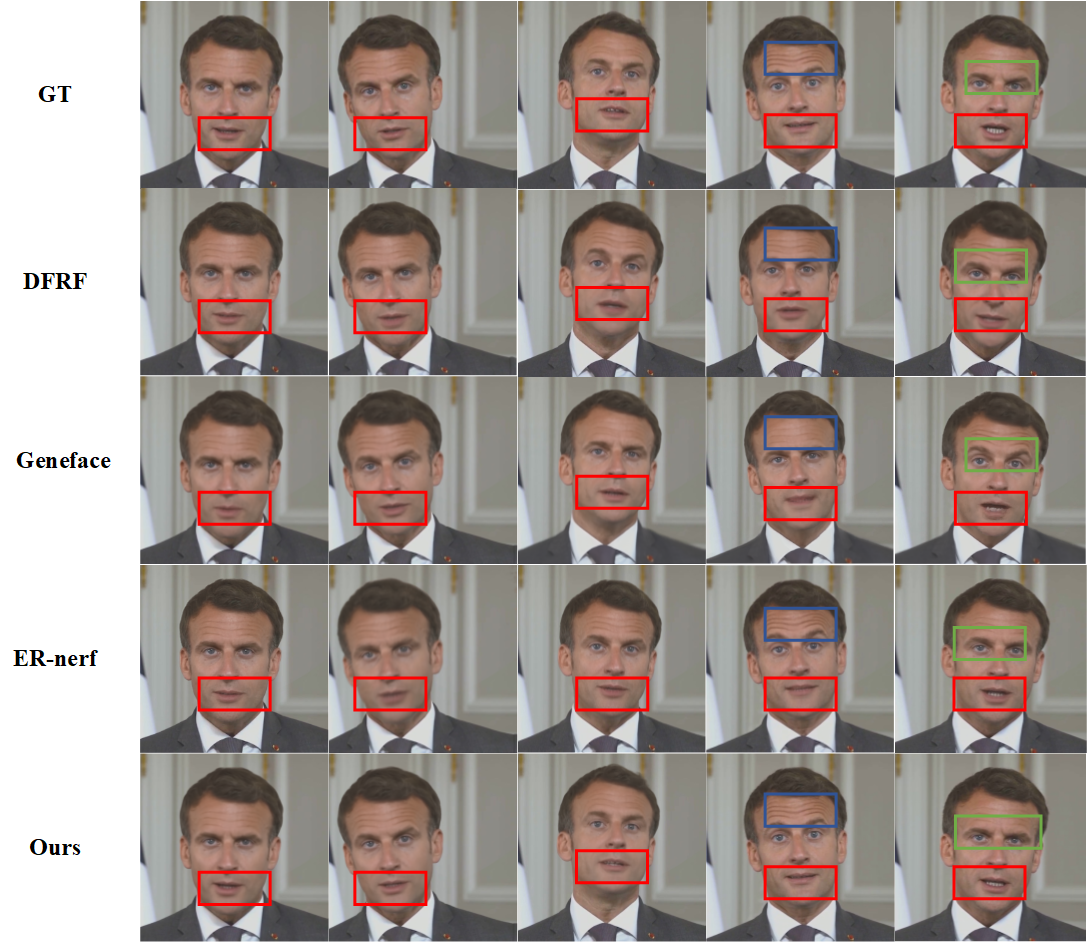}
\caption{ Illustration of the key frames and detailed comparisons of the generated portraits. Our method demonstrates significant advantages in overall visual coherence and lip synchronization on the Macron dataset.}
\label{fig}
\end{figure}

\begin{table}[htbp]
\caption{Table 1 presents the quantitative comparison results on the Macron dataset at a resolution of 512x512 pixels.}
\begin{center}
\begin{tabular}{lccccc}
\hline
Method   & PSNR↑           & LPIPS↓         & LMD↓           & FID↓            & Sync↑                                   \\ 
GT       & N/A             & 0              & 0              & 0               & N/A                                     \\ \hline
DFRF     & 30.863          & 0.101          & 3.173          & 15.076          & 4.314           \\
Geneface & 34.267          & \textbf{0.027} & 2.694          & 10.102          & 6.645           \\
Er-Nerf  & 35.100          & 0.028          & 2.815          & 10.275          & 6.532           \\
Ours     & \textbf{35.907} & 0.025          & \textbf{2.801} & \textbf{10.170} &\textbf{6.742} \\ \hline
\end{tabular}
\end{center}
\end{table}

To validate the generalizability of our method, comparative experiments have been conducted. We use collected videos of Obama\cite{guo2021AD-nerf} to conduct experiments under various scenarios. It is observed that our method demonstrates significant advantages in posture, eye status, and artifact removal.

\begin{figure}[htbp]
\centering
\includegraphics[width=0.5\textwidth]{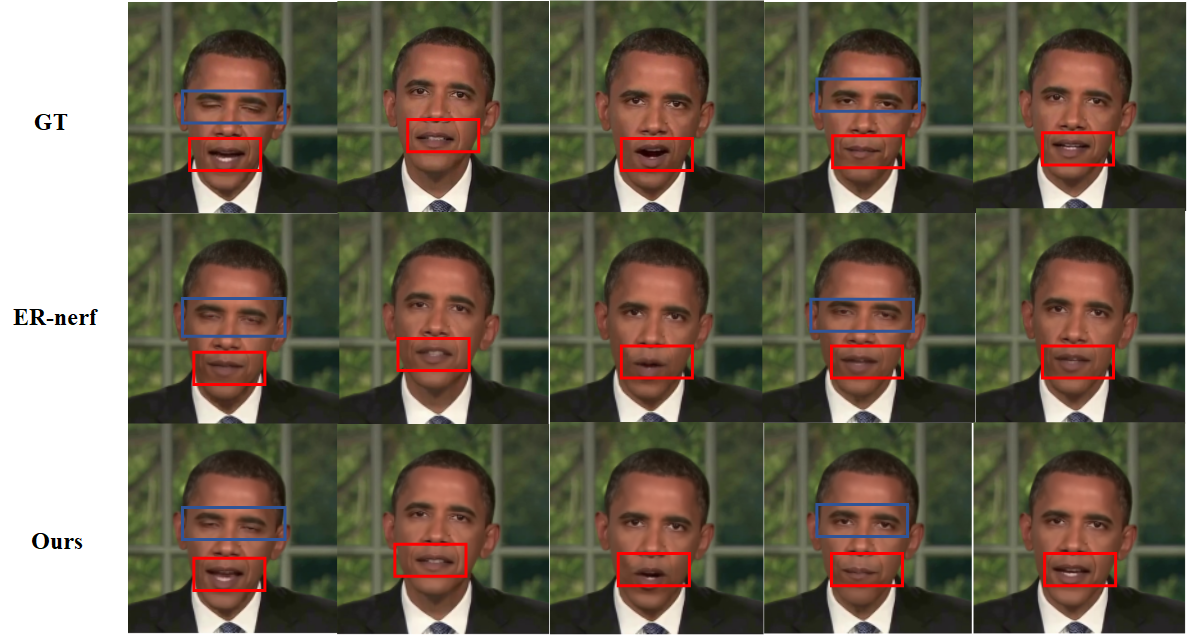}
\caption{Comparisons on the collected videos of Obama. Our method exhibits considerable advantages in mouth details, eye dynamics, and artifact reduction.}
\label{fig}
\end{figure}

\begin{table}[htbp]
\caption{Table 2 presents the quantitative comparisons of details conducted on the videos of Obama. The Er-nerf mouth change is smoother, so it has a higher score at LMD, but because of that we score better on Sync.}
\begin{center}
\begin{tabular}{lccccc}
\hline
Method  & PSNR↑           & LPIPS↓         & LMD↓           & FID↓           & Sync↑          \\
GT      & N/A             & 0              & 0              & 0              & N/A            \\ \hline
Er-Nerf & \textbf{37.353} & 0.021          & \textbf{2.492}          & 9.642          & 6.734          \\
Ours    & 36.754         & \textbf{0.016} & 2.584 & \textbf{9.431} & \textbf{6.908} \\ \hline
\end{tabular}
\end{center}
\end{table}

It can be observed that Er-nerf achieves higher scores in LMD due to its smoother variation in mouth shapes. However, our approach excels in predicting the corresponding mouth shapes for given audio inputs. Consequently, while our method may not achieve the same level of smoothness in mouth movements as Er-nerf, it results in lower LMD scores. Nonetheless, this discrepancy allows our approach to outperform Er-nerf in terms of Sync.

\subsection{ Ablation Study}

To validate the efficacy of our approach, we have conducted ablation studies for verification. When the dynamic landmark transformer is not used, the metrics are as follows. Additionally, a comparison is made in the absence of tri-plane hashing.

\begin{table}[htbp]
\caption{Table 2. Ablation Study Results on Macron's 512x512 Videos. D; dynamic landmark (enhanced sync), while E represents the eye movement generator.}
\begin{center}
\begin{tabular}{lccccc}

\hline
Methods            & PSNR↑           & LPIPS↓          & LMD↓           & SYNC↑          \\
GT                 & N/A             & 0               & 0              & N/A            \\ \hline
w/o D w/o E        & 35.438          & 0.0347          & 2.721          & 6.231          \\
w/o E w D          & \textbf{36.194} & 0.0274          & 2.731          & 6.685          \\
w E w D            & 35.891          & 0.0259          & 2.729          & 6.704          \\
w Tri-Hash w E w D & 35.907          & \textbf{0.0256} & \textbf{2.691} & \textbf{6.742} \\ \hline
\end{tabular}
\end{center}
\end{table}

\section{Ethical Consideration}
We specialize in creating high-quality talking-head videos, crucial for the emerging metaverse and digital human landscape. Our aim is to utilize this technology ethically, contributing positively to society while minimizing virtual impersonation and voice fraud. We anticipate our research to be referenced favorably, guiding the project towards lawful and ethical applications through deeper exploration. By advancing deep learning and digital human domains, we strive to foster positive progress in these fields.

\section{Conclusion}
In this paper, we propose an efficient framework for synthesizing realistic talking heads. By integrating audio features with dynamic facial landmark information, we successfully optimize the mouth region and enhance the representation capacity of the facial speaking area. The introduction of eye feature encoding further enriches facial features, resulting in more realistic synthesized heads. Utilizing triplane-hash and NeRF technologies, we effectively enhance the representation performance of facial regions, leading to more lifelike synthesized heads in three-dimensional space. Experimental results demonstrate significant achievements, with synthesized talking heads exhibiting a high degree of realism, and precise synchronization of mouth and eye movements with audio signals, thus yielding more natural and believable results.
Despite these accomplishments, two challenges persist. Firstly, synthesized faces still exhibit rigidity in their expressions. Secondly, specific training tailored to each target individual and scenario is required. Addressing these issues will be the focal point of our future work.


\bibliographystyle{IEEEtran}
\bibliography{ref}

\end{document}